\documentclass[11pt,a4paper]{article}
\usepackage[hyperref]{acl2019}
\usepackage{times}
\usepackage{latexsym}

\usepackage{url}

\aclfinalcopy 
\def\aclpaperid{2842} 

\usepackage{graphicx}
\usepackage{color, soul, multirow}
\usepackage{subfig}
\usepackage{bm}
\usepackage{verbatim}
\usepackage{amssymb}
\usepackage{pifont}
\usepackage{float}
\usepackage{amsmath}
\usepackage{booktabs}
\usepackage{pifont}
\usepackage{tikz}
\usepackage{pgfplots}
\newcommand{\cmark}{\ding{51}}%
\newcommand{\xmark}{\ding{55}}%

\title{Gated Embeddings in End-to-End Speech Recognition \\
for Conversational-Context Fusion}

\author{Suyoun Kim$^1$, Siddharth Dalmia$^2$ and Florian Metze$^2$ \\
  $^1$Electrical \& Computer Engineering\\
  $^2$Language Technologies Institute, School of Computer Science\\
  Carnegie Mellon University \\
  {\tt \{suyoung1, sdalmia, fmetze\}@andrew.cmu.edu} 
  }

\date{}

\begin{document}

\maketitle
\begin{abstract}
We present a novel conversational-context aware end-to-end speech recognizer based on a gated neural network that incorporates conversational-context/word/speech embeddings. Unlike conventional speech recognition models, our model learns longer conversational-context information that spans across sentences and is consequently better at recognizing long conversations. Specifically, we propose to use text-based external word and/or sentence embeddings (i.e., fastText, BERT) within an end-to-end framework, yielding significant improvement in word error rate with better conversational-context representation. We evaluated the models on the Switchboard conversational speech corpus and show that our model outperforms standard end-to-end speech recognition models. 
\end{abstract}

\section{Introduction}
In a long conversation, there exists a tendency of semantically related words, or phrases reoccur across sentences, or there exists topical coherence. Existing speech recognition systems are built at individual, isolated utterance level in order to make building systems computationally feasible. However, this may lose important conversational context information.  There have been many studies that have attempted to inject a longer context information \citep{mikolov2010recurrent, mikolov2012context, wang2015larger, ji2015document, liu2017dialog, xiong2018session}, all of these models are developed on text data for language modeling task.

There has been recent work attempted to use the conversational-context information within a \textit{end-to-end} speech recognition framework \citep{kim2018dialog, kimsituation, kim2019acoustic}. The new end-to-end speech recognition approach \citep{graves2006connectionist, graves2014towards, hannun2014deep, miao2015eesen, bahdanau2015neural, chorowski2015attention, chan2016listen, kim2017joint} integrates all available information within a single neural network model, allows to make fusing conversational-context information possible. However, these are limited to encode only one preceding utterance and learn from a few hundred hours of annotated speech corpus, leading to minimal improvements.

Meanwhile, neural language models, such as fastText \citep{bojanowski2017enriching, joulin2017bag, joulin2016fasttext}, ELMo \citep{peters2018deep}, OpenAI GPT \citep{radford2019language}, and Bidirectional Encoder Representations from Transformers (BERT) \citep{devlin2018bert}, that encode words and sentences in fixed-length dense vectors, embeddings, have achieved impressive results on various natural language processing tasks. Such general word/sentence embeddings learned on large text corpora (i.e., Wikipedia) has been used extensively and plugged in a variety of downstream tasks, such as question-answering and natural language inference, \cite{devlin2018bert, peters2018deep, Seo2017BidirectionalAF}, to drastically improve their performance in the form of transfer learning.

In this paper, we create a conversational-context aware end-to-end speech recognizer capable of incorporating a conversational-context to better process long conversations. Specifically, we propose to exploit external word and/or sentence embeddings which trained on massive amount of text resources, (i.e. fastText, BERT) so that the model can learn better conversational-context representations. So far, the use of such pre-trained embeddings have found limited success in the speech recognition task. We also add a gating mechanism to the decoder network that can integrate all the available embeddings (word, speech, conversational-context) efficiently with increase representational power using multiplicative interactions. Additionally, we explore a way to train our speech recognition model even with text-only data in the form of pre-training and joint-training approaches. We evaluate our model on the Switchboard conversational speech corpus \citep{swbd, godfrey1992switchboard}, and show that our model outperforms the sentence-level end-to-end speech recognition model. The main contributions of our work are as follows:
\begin{itemize}
    \item We introduce a contextual gating mechanism to incorporate multiple types of embeddings, word, speech, and conversational-context embeddings. 
    \item We exploit the external word (fastText) and/or sentence embeddings (BERT) for learning better conversational-context representation. 
    \item We perform an extensive analysis of ways to represent the conversational-context in terms of the number of utterance history, and sampling strategy considering to use the generated sentences or the true preceding utterance. 
    \item We explore a way to train the model jointly even with text-only dataset in addition to annotated speech data.
\end{itemize}

\section{Related work}
Several recent studies have considered to incorporate a context information within a end-to-end speech recognizer \citep{pundak2018deep,alon2019contextual}. In contrast with our method which uses a conversational-context information in a long conversation, their methods use a list of phrases (i.e. play a song) in reference transcription in specific tasks, \textit{contact names, songs names, voice search, dictation}. 

Several recent studies have considered to exploit a longer context information that spans multiple sentences \citep{mikolov2012context, wang2015larger, ji2015document, liu2017dialog, xiong2018session}. In contrast with our method which uses a single framework for speech recognition tasks, their methods have been developed on text data for language models, and therefore, it must be integrated with a conventional acoustic model which is built separately without a longer context information. 

Several recent studies have considered to embed a longer context information within a end-to-end framework \citep{kim2018dialog, kimsituation, kim2019acoustic}. In contrast with our method which can learn a better conversational-context representation with a gated network that incorporate external word/sentence embeddings from multiple preceding sentence history, their methods are limited to learn  conversational-context representation from one preceding sentence in annotated speech training set. 


Gating-based approaches have been used for fusing word embeddings with visual representations in genre classification task or image search task \citep{arevalo2017gated, kiros2018illustrative} and for learning different languages in speech recognition task \citep{kim2018towards}.

\section{End-to-End Speech Recognition Models}
\label{sec:review}
\subsection{Joint CTC/Attention-based encoder-decoder network}
We perform end-to-end speech recognition using a joint CTC/Attention-based approach with graphemes as the output symbols \cite{kim2017joint, watanabe2017hybrid}. The key advantage of the joint CTC/Attention framework is that it can address the weaknesses of the two main end-to-end models, Connectionist Temporal Classification (CTC) \cite{graves2006connectionist} and attention-based encoder-decoder (Attention) \cite{bahdanau2016end}, by combining the strengths of the two. With CTC, the neural network is trained according to a maximum-likelihood training criterion computed over all possible segmentations of the utterance's sequence of feature vectors to its sequence of labels while preserving left-right order between input and output. With attention-based encoder-decoder models, the decoder network can learn the language model jointly without relying on the conditional independent assumption.

Given a sequence of acoustic feature vectors, $\bm x$, and the corresponding graphemic label sequence, $\bm y$, the joint CTC/Attention objective is represented as follows by combining two objectives with a tunable parameter $\lambda: 0 \leq \lambda \leq 1$:
	\begin{align}
	    \label{eq:lambda}
		\mathcal{L} &= \lambda \mathcal{L}_\text{CTC} + (1-\lambda) \mathcal{L}_\text{att}.
	\end{align}
Each loss to be minimized is defined as the negative log likelihood of the ground truth character sequence $\bm{y^*}$, is computed from:
    \begin{align}
        \label{eq:loss_ctc}
        \begin{split}
        \mathcal{L}_\text{CTC} \triangleq & -\ln \sum_{\bm{\pi} \in \Phi(\bm{y})} p(\bm{\pi}|\bm{x}) 
        \end{split}
    \end{align}
    \begin{align}
        \label{eq:loss_att}
        \begin{split}
		\mathcal{L}_\text{att} \triangleq & -\sum_u \ln p(y_u^*|\bm{x},y^*_{1:u-1}) 
        \end{split}
    \end{align}
where $\bm{\pi}$ is the label sequence allowing the presence of the \textit{blank} symbol, $\Phi$ is the set of all possible $\bm{\pi}$ given $u$-length $\bm{y}$, and $y^*_{1:u-1}$ is all the previous labels. 

Both CTC and the attention-based encoder-decoder networks are also used in the inference step. The final hypothesis is a sequence that maximizes a weighted conditional probability of CTC and attention-based encoder-decoder network \cite{hori2017advances}:
    \begin{align}
        \label{eq:gamma}
        \begin{split}
        \bm{y}* = \text{argmax} \{ & \gamma \log p_{CTC}(\bm{y}|\bm{x}) \\
                      &+ (1-\gamma) \log p_{att}(\bm{y}|\bm{x}) \}
        \end{split}
    \end{align}

\subsection{Acoustic-to-Words Models}
\label{sec:a2w}
In this work, we use word units as our model outputs instead of sub-word units. Direct acoustics-to-word (A2W) models train a single neural network to directly recognize words from speech without any sub-word units, pronunciation model, decision tree, decoder, which significantly simplifies the training and decoding process \cite{soltau2016neural, Audhkhasi2017DirectAM, audhkhasi2018building, li2018advancing, palaskar2018acoustic}. In addition, building A2W can learn more semantically meaningful conversational-context representations and it allows to exploit external resources like word/sentence embeddings where the unit of representation is generally words. However, A2W models require more training data compared to conventional sub-word models because it needs sufficient acoustic training examples per word to train well and need to handle out-of-vocabulary(OOV) words. As a way to manage this OOV issue, we first restrict the vocabulary to 10k frequently occurring words. We then additionally use a single character unit and start-of-OOV \text{(sunk)}, end-of-OOV \text{(eunk)} tokens to make our model generate a character by decomposing the OOV word into a character sequence. For example, the OOV word, \textit{rainstorm}, is decomposed into (sunk) \textit{r a i n s t o r m} (eunk) and the model tries to learn such a character sequence rather than generate the OOV token. From this method, we obtained 1.2\% - 3.7\% word error rate (WER) relative improvements in evaluation set where exists 2.9\% of OOVs.

\section{Conversational-context Aware Models}
In this section, we describe the A2W model with conversational-context fusion. In order to fuse conversational context information within the A2W, end-to-end speech recognition framework, we extend the decoder sub-network to predict the output additionally conditioning on conversational context, by learning a conversational-context embedding. We encode single or multiple preceding utterance histories into a fixed-length, single vector, then inject it to the decoder network as an additional input at every output step. 

Let say we have $K$ number of utterances in a conversation. For $k$-th sentence, we have acoustic features $(x_1, \cdots, x_T)^k$ and output word sequence, $(w_1, \cdots, w_U)$. At output timestamp $u$, our decoder generates the probability distribution over words ($w_u^k$), conditioned on 1) speech embeddings, attended high-level representation ($\bm{e_{speech}^{k}}$) generated from encoder, and 2) word embeddings from all the words seen previously ($e^{u-1}_{word}$), and 3) conversational-context embeddings ($e^{k}_{context}$), which represents the conversational-context information for current ($k$) utterance prediction:
    \begin{align}
        \label{eq:loss}
        \bm{e^{k}_{speech}}       = & \text{Encoder}(\bm{x^k}) \\
        w^k_u \sim & \text{Decoder}(\bm{e^{k}_{context}}, e^k_{word}, \bm{e^{k}_{speech}})
    \end{align}    

We can simply represent such contextual embedding, $e^{k}_{context}$, by mean of one-hot word vectors or word distributions, $\texttt{mean}(e^{k-1}_{word_{1}} + \cdots + e^{k-1}_{word_{U}})$  from the preceding utterances. 

In order to learn and use the conversational-context during training and decoding, we serialize the
utterances based on their onset times and their conversations rather than random shuffling of data.
We shuffle data at the conversation level and create mini-batches that contain only one sentence of each conversation. We fill the "dummy" input/output example at positions where the conversation ended earlier than others within the mini-batch to not influence other conversations while passing context to the next batch.

\subsection{External word/sentence embeddings}

\begin{figure*}[t]
    \centering
    \centerline{\includegraphics[height=2in]{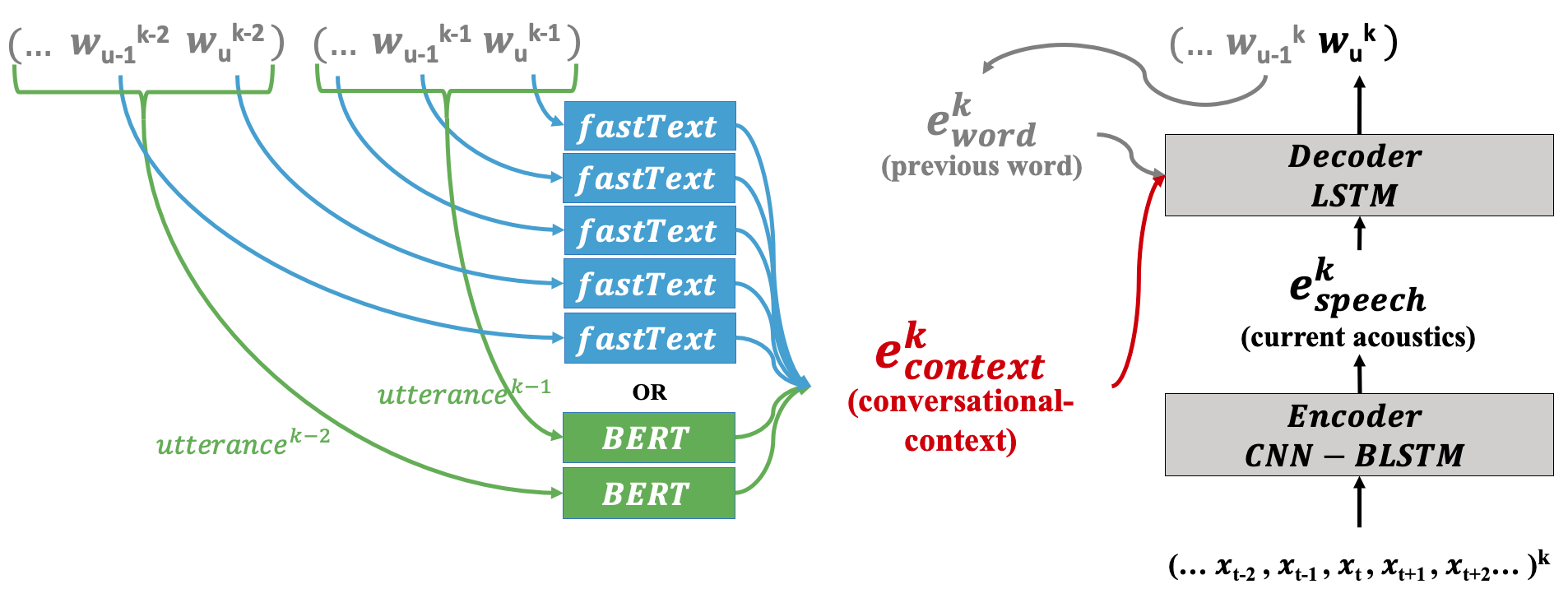}}
    \caption{Conversational-context embedding representations from external word or sentence embeddings.}
    \label{fig:context_embedding}
\end{figure*}
Learning better representation of conversational-context is the key to achieve better processing of long conversations. To do so, we propose to encode the general word/sentence embeddings pre-trained on large textual corpora within our end-to-end speech recognition framework. Another advantage of using pre-trained embedding models is that we do not need to back-propagate the gradients across contexts, making it easier and faster to update the parameters for learning a conversational-context representation. 

There exist many word/sentence embeddings which are publicly available. We can broadly classify them into two categories: (1) non-contextual word embeddings, and (2) contextual word embeddings. 
Non-contextual word embeddings, such as Word2Vec \cite{mikolov2012context}, GloVe \cite{pennington2014glove}, fastText \cite{bojanowski2017enriching}, maps each word independently on the context of the sentence where the word occur in. Although it is easy to use, it assumes that each word represents a single meaning which is not true in real-word. Contextualized word embeddings, sentence embeddings, such as deep contextualized word representations \cite{peters2018deep}, BERT \cite{devlin2018bert}, encode the complex characteristics and meanings of words in various context by jointly training a bidirectional language model. The BERT model proposed a masked language model training approach enabling them to also learn good ``sentence'' representation in order to predict the masked word.

In this work, we explore both types of embeddings to learn conversational-context embeddings as illustrated in Figure \ref{fig:context_embedding}. The first method is to use word embeddings, fastText, to generate 300-dimensional embeddings from 10k-dimensional one-hot vector or distribution over words of each previous word and then merge into a single context vector, $e^k_{context}$. Since we also consider multiple word/utterance history, we consider two simple ways to merge multiple embeddings (1) mean, and (2) concatenation. 
The second method is to use sentence embeddings, BERT. It is used to a generate single 786-dimensional sentence embedding from 10k-dimensional one-hot vector or distribution over previous words and then merge into a single context vector with two different merging methods. Since our A2W model uses a restricted vocabulary of 10k as our output units and which is different from the external embedding models, we need to handle out-of-vocabulary words. For fastText, words that are missing in the pretrained embeddings we map them to a random multivariate normal distribution with the mean as the sample mean and variance as the sample variance of the known words. For BERT, we use its provided tokenizer to generates byte pair encodings to handle OOV words. 

Using this approach, we can obtain a more dense, informative, fixed-length vectors to encode conversational-context information, $e^k_{context}$  to be used in next $k$-th utterance prediction. 

\subsection{Contextual gating}
We use contextual gating mechanism in our decoder network to combine the conversational-context embeddings with speech and word embeddings effectively. Our gating is contextual in the sense that multiple embeddings compute a gate value that is dependent on the context of multiple utterances that occur in a conversation. Using these contextual gates can be beneficial to decide how to weigh the different embeddings, conversational-context, word and speech embeddings. Rather than merely concatenating conversational-context embeddings \cite{kim2018dialog}, contextual gating can achieve more improvement because its increased representational power using multiplicative interactions. 

\begin{figure}[!h]
    \begin{minipage}[b]{1.0\linewidth}
        \centering
        \centerline{\includegraphics[width=8.5cm]{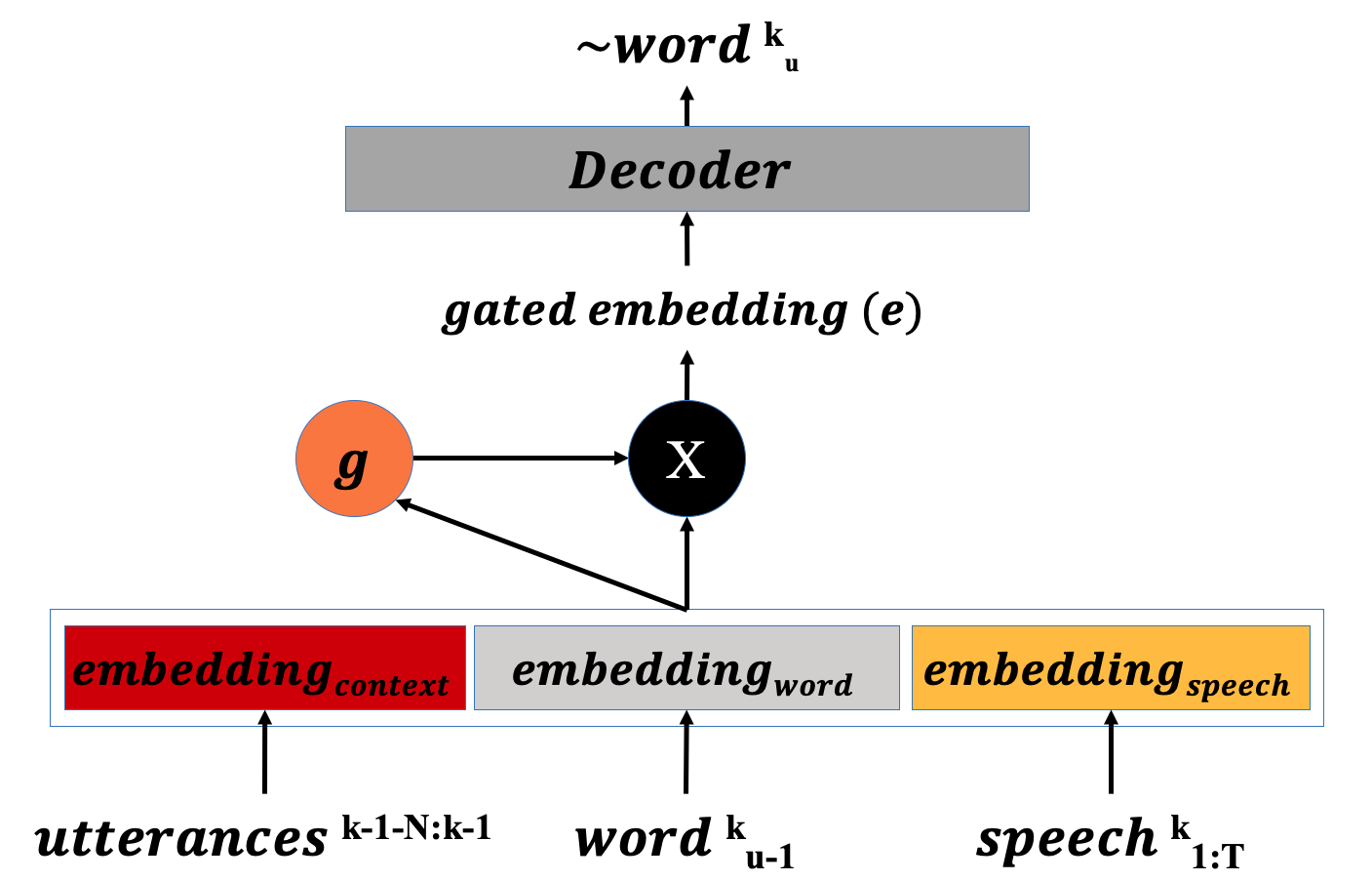}}
    \end{minipage}
    \caption{Our contextual gating mechanism in decoder network to integrate three different embeddings from: 1) conversational-context, 2) previous word, 3) current speech. }
    \label{fig:gate}
\end{figure}

Figure \ref{fig:gate} illustrates our proposed contextual gating mechanism. Let $e_w = e_w(y_{u-1})$ be our previous word embedding for a word $y_{u-1}$, and let $e_s = e_s(x^k_{1:T})$ be a speech embedding for the acoustic features of current $k$-th utterance $x^k_{1:T}$ and $e_c = e_c(s_{k-1-n:k-1})$ be our conversational-context embedding for $n$-number of preceding utterances ${s_{k-1-n:k-1}}$. Then using a gating mechanism:
\begin{align}
    g = \sigma (e_c, e_w, e_s) 
\end{align}
where $\sigma$ is a 1 hidden layer DNN with $\texttt{sigmoid}$ activation, the gated embedding $e$ is calcuated as
\begin{align}
    e = g \odot (e_c, e_w, e_s) \\
    h = \text{LSTM}(e)
\end{align}
and fed into the LSTM decoder hidden layer. The output of the decoder $h$ is then combined with conversational-context embedding $e_c$ again with a gating mechanism,
\begin{align}
    g = \sigma (e_C, h) \\
    \hat h = g \odot (e_c, h)
\end{align}
Then the next hidden layer takes these gated activations, $\hat h$, and so on. 

\section{Experiments}
\label{sec:exp}
\subsection{Datasets}

To evaluate our proposed  conversational end-to-end speech recognition model, we use the Switchboard (SWBD) LDC corpus (97S62) task. We split 300 hours of the SWBD training set into two: 285 hours of data for the model training, and 5 hours of data for the hyper-parameter tuning. We evaluate the model performance on the HUB5 Eval2000 which consists of the Callhome English (CH) and Switchboard (SWBD) (LDC2002S09, LDC2002T43). In Table \ref{tab:data}, we show the number of conversations and the average number of utterances per a single conversation. 

\begin{table}[t]
\begin{center}
\resizebox{\columnwidth}{!}{
\begin{tabular}{r|r|r|r}
\toprule
Dataset & \# of utter. & \# of conversations & avg. \# of utter.\\
       &        &             & $/$conversation \\       
\midrule
training   & 192,656 & 2402 & 80\\
validation    & 4,000 & 34 & 118 \\
eval.(SWBD) & 1,831 & 20 & 92\\
eval.(CH) & 2,627 & 20 & 131 \\
\bottomrule
\end{tabular}
}
\end{center}
\caption{ Experimental dataset description. We used 300 hours of Switchboard conversational corpus. Note that any pronunciation lexicon or Fisher transcription was not used.}
\label{tab:data}
\end{table}

The audio data is sampled at 16kHz, and then each frame is converted to a 83-dimensional feature vector consisting of 80-dimensional log-mel filterbank coefficients and 3-dimensional pitch features as suggested in \cite{miao2016empirical}. The number of our word-level output tokens is 10,038, which includes 47 single character units as described in Section \ref{sec:a2w}. Note that no pronunciation lexicon was used in any of the experiments.

\begin{table*}[t]
\centering
\resizebox{\textwidth}{!}{
\begin{tabular}{r|r|r|r|r|r}
\toprule
   &    & Trainable & External & SWBD & CH \\
Model   & Output Units & Params  & LM & (WER\%) & (WER\%) \\
\midrule
\textbf{Prior Models} & & & & & \\
LF-MMI \citep{povey2016purely} & CD phones & N/A & \cmark & 9.6 & 19.3 \\
CTC \citep{zweig2017advances} & Char & 53M & \cmark & 19.8 & 32.1 \\
CTC \citep{sanabria2018hierarchical} & Char, BPE-\{300,1k,10k\} & 26M & \cmark & 12.5 &23.7 \\
CTC \citep{audhkhasi2018building} & Word (Phone init.) & N/A & \cmark & 14.6  & 23.6 \\
Seq2Seq \citep{zeyer2018improved} & BPE-10k & 150M* & \xmark & 13.5 & 27.1 \\
Seq2Seq \citep{palaskar2018acoustic} & Word-10k & N/A & \xmark & 23.0 & 37.2 \\
Seq2Seq \citep{zeyer2018improved} & BPE-1k & 150M* & \cmark & 11.8 & 25.7 \\
\midrule
\textbf{Our baseline} & Word-10k & 32M & \xmark & 18.2 & 30.7 \\
\midrule
\textbf{Our Proposed Conversational Model} & & & & & \\
Gated Contextual Decoder & Word-10k & 35M & \xmark & 17.3 & 30.5  \\
+ Decoder Pretrain & Word-10k & 35M & \xmark & 16.4 & 29.5  \\
+ fastText for Word Emb. & Word-10k & 35M & \xmark & 16.0 & 29.5 \\
(a) fastText for Conversational Emb. & Word-10k & 34M & \xmark & 16.0 & 29.5 \\
(b) BERT for Conversational Emb. & Word-10k & 34M & \xmark & 15.7 & 29.2 \\
(b) + Turn number 5 & Word-10k & 34M & \xmark & \textbf{15.5} & \textbf{29.0} \\
\bottomrule
\end{tabular}
}
\caption{Comparison of word error rates (WER) on Switchboard 300h with standard end-to-end speech recognition models and our proposed end-to-end speech recogntion models with conversational context. (The * mark denotes our estimate for the number of parameters used in the previous work).}
\label{tab:result}
\end{table*}

\subsection{Training and decoding}
\label{sec:training}
For the architecture of the end-to-end speech recognition, we used joint CTC/Attention end-to-end speech recognition \citep{kim2017joint, watanabe2017hybrid}. As suggested in \citep{zhang2017very, hori2017advances}, the input feature images are reduced to ($1/4 \times 1/4$) images along with the time-frequency axis within the two max-pooling layers in CNN. Then, the 6-layer BLSTM with 320 cells is followed by the CNN layer. For the attention mechanism, we used a location-based method \citep{chorowski2015attention}. For the decoder network, we used a 2-layer LSTM with 300 cells. In addition to the standard decoder network, our proposed models additionally require extra parameters for gating layers in order to fuse conversational-context embedding to the decoder network compared to baseline. We denote the total number of trainable parameters in Table \ref{tab:result}. 

For the optimization method, we use AdaDelta \citep{zeiler2012adadelta} with gradient clipping \citep{pascanu2013difficulty}. We used $\lambda = 0.2$ for joint CTC/Attention training (in Eq. \ref{eq:lambda}) and $\gamma = 0.3$ for joint CTC/Attention decoding (in Eq.\ref{eq:gamma}). We bootstrap the training of our proposed conversational end-to-end models from the baseline end-to-end models. To decide the best models for testing, we monitor the development accuracy where we always use the model prediction in order to simulate the testing scenario. At inference, we used a left-right beam search method \citep{sutskever2014sequence} with the beam size 10 for reducing the computational cost. We adjusted the final score, $s(\bm{y}|\bm{x})$, with the length penalty $0.5$. The models are implemented using the PyTorch deep learning library \citep{paszke2017automatic}, and ESPnet toolkit \citep{kim2017joint, watanabe2017hybrid, watanabe2018espnet}.

\section{Results}
\label{sec:result}
Our results are summarized in the Table \ref{tab:result} where we first present the baseline results and then show the improvements by adding each of the individual components that we discussed in previous sections, namely, gated decoding, pretraining decoder network, external word embedding, external conversational embedding and increasing receptive field of the conversational context. Our best model gets around 15\% relative improvement on the SWBD subset and 5\% relative improvement on the CallHome subset of the eval2000 dataset. 

We start by evaluating our proposed model which leveraged conversational-context embeddings learned from training corpus and compare it with a standard end-to-end speech recognition models without conversational-context embedding. As seen in Table \ref{tab:result}, we obtained a performance gain over the baseline by using conversational-context embeddings which is learned from training set. 

\subsection{Pre-training decoder network}
Then, we observe that pre-training of decoder network can improve accuracy further as shown in Table \ref{tab:result}. Using pre-training the decoder network, we achieved 5\% relative improvement in WER on SWBD set. Since we add external parameters in decoder network to learn conversational-context embeddings, our model requires more efforts to learn these additional parameters. To relieve this issue, we used pre-training techniques to train decoder network with text-only data first. We simply used a mask on top of the Encoder/Attention layer so that we can control the gradients of batches contains text-only data and do not update the Encoder/Attention sub-network parameters. 

\subsection{Use of words/sentence embeddings}
Next, we evaluated the use of pretrained external embeddings (fastText and BERT). We initially observed that we can obtain 2.4\% relative improvement over (the model with decoder pretraining) in WER by using fastText for additional word embeddings to the gated decoder network. 

We also extensively evaluated various ways to use fastText/BERT for conversational-context embeddings. Both methods with fastText and with BERT shows significant improvement from the baseline as well as vanilla conversational-context aware model.

\subsection{Conversational-context Receptive Field}
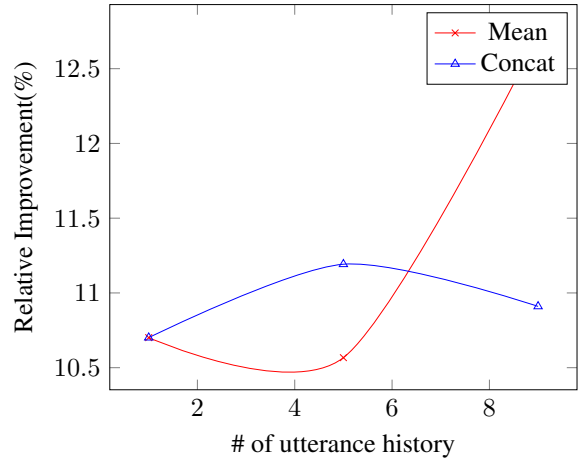
\begin{figure}[H]
\resizebox {\columnwidth} {!} {

\begin{tikzpicture}
	\begin{axis}[
		xlabel=\# of utterance history,
		ylabel=Relative Improvement(\%) ]
	\addplot[smooth,color=red,mark=x] coordinates {
	    (1,  10.701)
	    (5,  10.567)
	    (9,  12.714)
	};
	\addplot[smooth,color=blue,mark=triangle] coordinates {
	    (1,  10.701)
	    (5,  11.192)
	    (9,  10.910)
	};
	\legend{Mean, Concat}

	\end{axis}
\end{tikzpicture}
}
    \caption{The relative improvement in Development accuracy over sets over baseline obtained by using conversational-context embeddings with different number of utterance history and different merging techniques.}
    \label{fig:turn_number}
\end{figure}

We also investigate the effect of the number of utterance history being encoded. We tried different $N = [1, 5, 9]$ number of utterance histories to learn the conversational-context embeddings. Figure \ref{fig:turn_number} shows the relative improvements in the accuracy on the Dev set (\ref{sec:training}) over the baseline ``non-conversational'' model. We show the improvements on the two different methods of merging the contextual embeddings, namely mean and concatenation. 
Typically increasing the receptive field of the conversational-context helps improve the model. However, as the number of utterence history increased, the number of trainable parameters of the concatenate model increased making it harder for the model to train. This led to a reduction in the accuracy.

We also found that using 5-utterance history with concatenation performed best (15\%) on the SWBD set, and using 9-number of utterance history  with mean method performed best (5\%) on CH set. We also observed that the improvement diminished when we used 9-utterance history for SWBD set, unlike CH set. One possible explanation is that the conversational-context may not be relevant to the current utterance prediction or the model is overfitting.  

\subsection{Sampling technique}
\begin{figure}[H]
\resizebox {\columnwidth} {!} {

\begin{tikzpicture}
	\begin{axis}[
		xlabel=Utterance Sampling Rate(\%),
		ylabel=Relative Improvement(\%) ]
	\addplot[smooth,color=red,mark=x] coordinates {
	    (0,   1.175)
	    (20,  3.247)
	    (50,  2.755)
	    (100, 0)
	};
	\legend{Accuracy on Dev. set}

	\end{axis}
\end{tikzpicture}
}
    \caption{The relative improvement in Development accuracy over 100\% sampling rate which was used in \cite{kim2018dialog} obtained by using conversational-context embeddings with different sampling rate.}
\label{fig:sampling}

\end{figure}
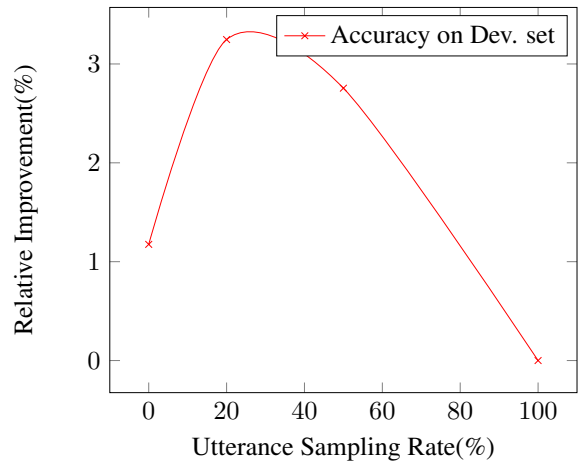
We also experiment with an utterance level sampling strategy with various sampling ratio, $[0.0, 0.2, 0.5, 1.0]$. Sampling techniques have been extensively used in sequence prediction tasks to reduce overfitting \citep{bengio2015scheduled} by training the model conditioning on generated tokens from the model itself, which is how the model actually do at inference, rather than the ground-truth tokens. Similar to choosing previous word tokens from the ground truth or from the model output, we apply it to choose previous utterance from the ground truth or from the model output for learning conversational-context embeddings. Figure \ref{fig:sampling} shows the relative improvement in the development accuracy (\ref{sec:training}) over the $1.0$ sampling rate which is always choosing model's output. We found that a sampling rate of 20\% performed best. 

\subsection{Analysis of context embeddings}

We develop a scoring function, $s(i,j)$ to check if our model conserves the conversational consistency for validating the accuracy improvement of our approach. The scoring function measures the average of the conversational distances over every consecutive hypotheses generated from a particular model. The conversational distance is calculated by the Euclidean distance, $\text{dist}(e_i, e_j)$ of the fixed-length vectors $e_i, e_j$ which represent the model's $i, j$-th hypothesis, respectively. To obtain a fixed-length vector, utterance embedding, given the model hypothesis, we use BERT sentence embedding as an oracle. Mathematically it can be written as,
\begin{align*}
    s(i,j) = \frac{1}{N}\sum_{i,j \in \texttt{eval}}(\text{dist}(e_i,e_j))
\end{align*}
where, $i, j$ is a pair of consecutive hypotheses in evaluation data $\texttt{eval}$, $N$ is the total number of $i,j$ pairs, $e_i, e_j$ are BERT embeddings. In our experiment, we select the pairs of consecutive utterances from the reference that show lower distance score at least baseline hypotheses. 

\begin{figure}[!h]
    \begin{minipage}[b]{1.0\linewidth}
        \centering
        \centerline{\includegraphics[width=7cm]{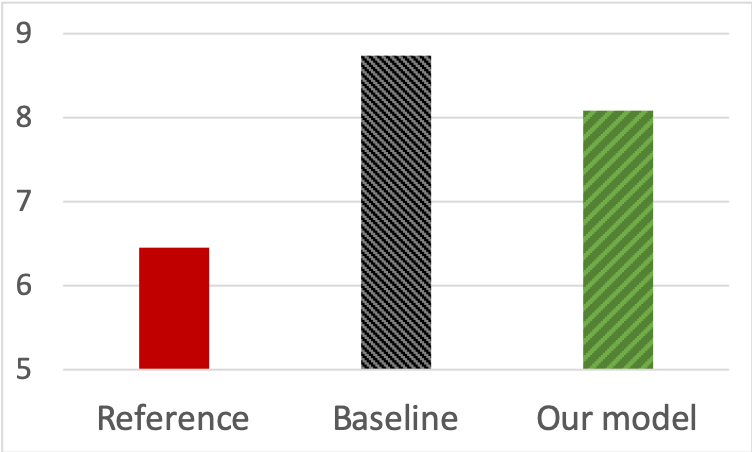}}
    \end{minipage}
    \caption{Comparison of the conversational distance score on the consecutive utterances of 1) reference, 2) our proposed conversational end-to-end model, and 3) our end-to-end baseline model.}
    \label{fig:score}
\end{figure}

From this process, we obtained three conversational distance scores from 1) the reference transcripts, 2) the hypotheses of our vanilla conversational model which is not using BERT, and 3) the hypotheses of our baseline model. Figure \ref{fig:score} shows the score comparison.

We found that our proposed model was 7.4\% relatively closer to the reference than the baseline. This indicates that our conversational-context embedding leads to improved similarity across adjacent utterances, resulting in better processing a long conversation.

\section{Conclusion}
\label{sec:conclusion}

We have introduced a novel method for conversational-context aware end-to-end speech recognition based on a gated network that incorporates word/sentence/speech embeddings. Unlike prior work, our model is trained on conversational datasets to predict a word, conditioning on multiple preceding conversational-context representations, and consequently improves recognition accuracy of a long conversation. Moreover, our gated network can incorporate effectively with text-based external resources, word or sentence embeddings (i.e., fasttext, BERT) within an end-to-end framework and so that the whole system can be optimized towards our final objectives, speech recognition accuracy. By incorporating external embeddings with gating mechanism, our model can achieve further improvement with better conversational-context representation. We evaluated the models on the Switchboard conversational speech corpus and show that our proposed model using gated conversational-context embedding show 15\%, 5\% relative improvement in WER compared to a baseline model for Switchboard and CallHome subsets respectively. Our model was shown to outperform standard end-to-end speech recognition models trained on isolated sentences. This work is easy to scale and can potentially be applied to any speech related task that can benefit from longer context information, such as spoken dialog system, sentimental analysis.

\section*{Acknowledgments}
We gratefully acknowledge the support of NVIDIA Corporation with the donation of the Titan Xp GPU used for this research. This work also used the Bridges system, which is supported by NSF award number ACI-1445606, at the Pittsburgh Supercomputing Center (PSC). 

\bibliography{acl2019}
\bibliographystyle{acl_natbib}

\appendix

\end{document}